\documentclass{article} 
\usepackage{collas2023_conference,times}
\usepackage{easyReview}

\usepackage{amsmath,amsfonts,bm}

\def\eqref#1{equation~\ref{#1}}

\def\1{\bm{1}}

\DeclareMathAlphabet{\mathsfit}{\encodingdefault}{\sfdefault}{m}{sl}
\SetMathAlphabet{\mathsfit}{bold}{\encodingdefault}{\sfdefault}{bx}{n}

\usepackage{hyperref}
\usepackage{algorithm}
\usepackage{algpseudocode} 
\hypersetup{
    colorlinks=true,
    linkcolor=red,
    filecolor=magenta,
    urlcolor=blue,
    citecolor=purple,
    pdftitle={Overleaf Example},
    pdfpagemode=FullScreen,
    }

\title{Auxiliary task discovery through generate-and-test}

\author{Banafsheh Rafiee  \\
Department of Computing Science\\
University of Alberta\\
Canada \\
\texttt{\{rafiee\}@ualberta.ca} \\
\And 
Sina Ghiassian  \\
Department of Computing Science\\
University of Alberta\\
Canada \\
\And 
Jun Jin \\
Huawei Technologies Canada, Ltd \\
Canada \\
\And 
Richard Sutton  \\
Department of Computing Science\\
University of Alberta\\
Canada \\
\And 
Jun Luo \\
Huawei Technologies Canada, Ltd \\
Canada \\
\And 
Adam White  \\
Department of Computing Science\\
University of Alberta\\
Canada \\
}
\collasfinalcopy 

\begin{document}

\maketitle

\begin{abstract}
In this paper, we explore an approach to auxiliary task discovery in reinforcement learning based on ideas from representation learning. Auxiliary tasks tend to improve data efficiency by forcing the agent to learn auxiliary prediction and control objectives in addition to the main task of maximizing reward, and thus producing better representations. Typically these tasks are designed by people. Meta-learning offers a promising avenue for automatic task discovery; however, these methods are computationally expensive and challenging to tune in practice. In this paper, we explore a complementary approach to the auxiliary task discovery: continually generating new auxiliary tasks and preserving only those with high utility. We also introduce a new measure of auxiliary tasks' usefulness based on how useful the features induced by them are for the main task. Our discovery algorithm significantly outperforms random tasks and learning without auxiliary tasks across a suite of environments. 
\end{abstract}

\section{Introduction}
The discovery question---what should an agent learn about---remains an open challenge for AI research. In the context of reinforcement learning, multiple components define the scope of what the agent is learning about. 
The agent's behavior defines its focus and attention in terms of data collection. 
Related exploration methods based on intrinsic rewards define what the agent chooses to do outside of reward maximization. 
Most directly, the auxiliary learning objectives we build in, including macro actions or options, models, and representation learning objectives force the agent to learn about other things beyond a reward maximizing policy. 
The primary question is where do these auxiliary learning objectives come from?

Classically, there are two approaches to defining auxiliary objectives that are the extremes of a spectrum of possibilities. 
The most common approach is for people to build the auxiliary objectives in pre-defining option policies, intrinsic rewards, and model learning objectives. 
Although most empirically successful, this approach has obvious limitations like feature engineering of old. At the other extreme is {\em end-to-end learning}. 
The idea is to build in as little inductive bias as possible including the inductive biases introduced by auxiliary learning objectives. Instead, we let the agent's neural network discover and adapt internal representations and algorithmic components (e.g., discovering objectives \citep{xu2020meta}, update rules \citep{oh2020discovering}, and models \citep{silver2017predictron}) just through trial and error interaction with the world. This approach remains challenging due to data efficiency concerns and in some cases shifts the difficulty from auxiliary objective design to loss function and curriculum design. 

An alternative approach that exists somewhere between human design and end-to-end learning is to hand-design many tasks in the form of additional output heads on the network that must be optimized in addition to the primary learning signal.
These tasks, called {\em auxiliary tasks}, exert pressure on the lower layers of the neural network during training, yielding agents that can learn faster \citep{mirowski2016learning,shelhamer2016loss}, produce better final performance \citep{jaderberg2016reinforcement}, and at times transfer to other related problems \citep{wang2022investigating}. 
This positive influence on neural network training is called the {\em auxiliary task effect} and is related to the emergence of good internal representations we seek in end-to-end learning. 
The major weakness of auxiliary task learning is its dependence on people. 
Relying on people for designing auxiliary tasks is not ideal because it is challenging to know what auxiliary tasks will be useful in advance and, as we will show later, poorly specified auxiliary tasks can significantly slow learning.

There has been relatively little work on autonomously discovering auxiliary tasks. One approach is to use meta learning. 
Meta-learning methods are higher-level learning methods that adapt the parameters of the base learning system, such as step-sizes, through gradient descent \citep{xu2018meta}. This approach can be applied to learning auxiliary tasks defined via General Value Functions or GVFs \citep{sutton2011horde} by adapting the parameters that define the goal (cumulant) and termination functions via gradient-descent \citep{veeriah2019discovery}. Generally speaking, these meta-learning approaches require large amounts of training data and are notoriously difficult to tune \citep{antoniou2018train}. 

An exciting alternative is to augment these meta-learning approaches with generate-and-test mechanisms that can discover new auxiliary tasks, which can later be refined via meta-learning. 
This approach has produced promising results in representation learning where simple generate-and-test significantly improves classification and regression performance when combined with back-prop \citep{dohare2021continual}. 
Before we can combine meta-learning and generate-and-test, we must first develop the generate-and-test approach to auxiliary task discovery so that their combination has the best chance for success. 
Such an effort is worthy of an entire study on its own, so in this paper we leave combining the two to future work and focus on the generate-and-test approach.

Despite significant interest, it remains unclear what makes a good or bad auxiliary tasks. The meta-learning approaches do not generate human-interpretable tasks. Updating toward multiple previous policies, called the {\em value improvement path} \citep{dabney2020value}, can improve performance but is limited to historical tasks. 
The gradient alignment between auxiliary tasks and the main task has been proposed as a measure of auxiliary tasks usefulness \citep{lin2019adaptive, du2018adapting}. However, the efficacy of this measure has not been thoroughly studied. 
Randomly generated auxiliary tasks can help avoid representation collapse \citep{lyle2021effect} and improve performance \citep{zheng2021learning}, but can also generate significant interference which degrades performance \citep{wang2022investigating}.   

In this paper we take a step toward understanding what makes useful auxiliary tasks, introducing a new generate-and-test method for autonomously generating new auxiliary tasks and a new measure of task usefulness to prune away bad ones.  
The proposed measure of task usefulness evaluates the auxiliary tasks based on how useful the features induced by them are for the main task.
Our experimental results shows that our measure of task usefulness successfully distinguishes between the good and bad auxiliary tasks.
Moreover, our proposed generate-and-test method outperforms random tasks and learning without auxiliary tasks.

\section{Background}
\label{sec:background}
In this paper, we consider the interaction of an agent with its environment at discrete time steps $t=1,2,\hdots$. The current state is denoted by $S_t \in \mathcal{S}$. The agent's action $A_t \in \mathcal{A}$ is selected according to a policy $\pi: \mathcal{A}\times\mathcal{S} \rightarrow [0, 1]$, causing the environment to transition to the next state $S_{t+1}$ emitting a reward of $R_{t+1} \in \mathbb{R}$. 
The goal of the agent is to find the policy $\pi$ with the highest state-action value function defined as $q_{\pi}(s, a) \doteq \mathbf{E}_\pi[G_t | S_t=s, A_t = a]$ where $G_t \doteq \sum_{k=0}^{\infty}\gamma^k R_{t+k+1}$ is called the return with $\gamma \in [0, 1)$ being the discount factor.

To estimate the state-action value function, we use temporal-difference learning \citep{sutton1988learning}. 
Specifically, we use Q-learning \citep{watkins1992q} to learn a parametric approximation $\hat{q}(s, a; \bf{w})$ by updating a vector of parameters ${\bf{w}} \in \mathbf{R}^d$. The update is as follows,
$$    {\bf{w_{t+1}}} \leftarrow {\bf{w_t}} + \alpha \delta_t \nabla_{\bf{w}_t} \hat{q}(S_t, A_t; \bf{w}_t)
,$$

\noindent where $\delta_t \doteq R_{t+1} + \gamma\mbox{max}_a \hat{q}(S_{t+1}, a; {\bf{w_t}}) - \hat{q}(S_t, A_t; \bf{w}_t)$ is the TD error,
 $\nabla_{\bf{w}_t} \hat{q}(S_t, A_t; \bf{w}_t)$ is the gradient of the value function with respect to the parameters $\bf{w}_t$, and 
the scalar $\alpha$ denotes the step-size parameter.
For action selection, Q-learning is commonly combined with an epsilon greedy policy.

We use neural networks for function approximation. We integrate a replay buffer, a target network, and the Adam optimizer with Q-learning as is commonly done to improve performance \citep{mnih2013playing}. 

To formulate auxiliary tasks, a common approach is to use general value functions or GVFs \citep{sutton2011horde}. GVFs are value functions with a generalized notion of target and termination. More specifically, a GVF can be written as the expectation of the discounted sum of any signal of interest: 
$$v_{\pi, \gamma, c}(s) \doteq \mathbf{E}_{\pi}[\sum_{k=0}^{\infty}(\prod_{j=1}^k\gamma(S_{t+j}))c(S_{t+k+1}) | S_t=s, A_{t:\infty} ~ \pi]
$$
\noindent where $\pi$ is the policy, $\gamma$ is the continuation function, and $c$ is a signal of interest and is referred to as the cumulant. Similarly, a generalized state-action value function $q_{\pi, \gamma, c}(s, a)$ can be defined where the expectation is conditioned on $A_t = a$ as well as $S_t = s$.
A control auxiliary tasks is one where the agent attempts to learn a $\pi$ to maximize the expected discounted sum of the future signal of interest (called a control demon or control GVF in prior work). 

To learn these auxiliary tasks, multi-headed neural networks are commonly used where the last hidden layer acts as the representation shared between the main task and the auxiliary tasks \citep{jaderberg2016reinforcement}.
In this setting, each head corresponds to either the main task or one of the auxiliary tasks and the auxiliary tasks make changes to the representation alongside the main task via backpropagation. 

\section{Auxiliary task discovery through generate-and-test}
We propose a new method for auxiliary task discovery based on a class of algorithms called generate-and-test. 
Generate-and-test was originally proposed as an approach to representation learning or feature finding. 
We can think of backprop with a large neural network as performing a massive parallel search in feature space \citep{frankle2018lottery}. 
Backprop greatly depends on the randomness in the weight initialization to find good features.
The idea of generate and test is to continually inject randomness in the feature search by continually proposing new features using a \textit{generator}, to measure features usefulness using a \textit{tester}, and to discard useless features.
This idea has a long history in supervised learning \citep{sutton2014online, mahmood2013representation}, and can even be combined with backprop \citep{dohare2021continual}.
The same basic structure can be applied to auxiliary task discovery, which we explain next.

We use generate-and-test for discovering and retaining auxiliary tasks that induce a representation useful for learning the main task. That is, the goal is to find auxiliary tasks that induce a positive auxiliary task effect.
It is challenging to recognize which auxiliary tasks induce useful representations.
To do so, we first evaluate how good each feature is based on how much it contributes to the approximation of the main task action-value function. 
Here we define the features to be the output of neural network's last hidden layer after applying the activation function. 
We then identify which auxiliary task was responsible for shaping which features. 


Our proposed generate-and-test method for discovering auxiliary tasks consists of a generator and a tester. 
The \textit{generator} generates new auxiliary tasks and the \textit{tester} evaluates the auxiliary tasks.
The auxiliary tasks that are assessed as useful are retained while the auxiliary tasks that are assessed to be useless are replaced by newly generated auxiliary tasks. 
The newly generated auxiliary tasks will most likely have low utility. 
To prevent the replacement of newly generated auxiliary tasks, we calculate the number of steps since their generation and refer to that as their age. 
An auxiliary task can only be replaced if its age is bigger than some age threshold. 
Every $T$ time step, some ratio of the auxiliary tasks get replaced. 
We refer to $T$ as the replacement cycle and denote the replacement ratio by $\rho$.
The pseudo-code for the proposed generate-and-test method is shown in Algorithm \ref{alg:alg}. 

\begin{algorithm}{}
	\caption{Generate-and-test for auxiliary task discovery} 
	\label{alg:alg}
	\begin{algorithmic}[1]
	\State \textbf{Input}: number of auxiliary tasks $n$, age threshold $\mu$, replacement cycle $T$, replacement ratio $\rho$
	\State \textbf{Initialization:}
	\State generate $n$ auxiliary tasks using the \textit{generator}
	\State randomly initialize the base learning network
 	\State set age $a_i$ for each auxiliary task to zero
	\For {Every time step}
    	\State do a DQN step to update the base learning network
     	\State Increase $a_i$ by one for $i=1, \dots, n$
    	\State update the utility of each auxiliary task $u(\mbox{aux}^i)$ for $i=1, ..., n$ using the \textit{tester}
    	\For {Every $T$ time steps}
    	\State Find $n\rho$ auxiliary tasks with the lowest utilities such that $a_i > \mu$
    	\State replace the $n\rho$ auxiliary tasks with new auxiliary tasks generated by the \textit{generator}
    	\State reinitialize the input and output weights of the features induced by the $n\rho$ auxiliary tasks
     	\State reset $a_i$ to zero for the $n\rho$ auxiliary tasks
            \EndFor
        \EndFor
	\end{algorithmic} 
\end{algorithm}

Note that the proposed method does not generate-and-test on features but on auxiliary tasks.
It, however, does assess the utility of features and derives the utility of the auxiliary tasks from the utility of the features that they induced.

\begin{figure}[!htbp]
\centering
  \includegraphics[width=1\linewidth]{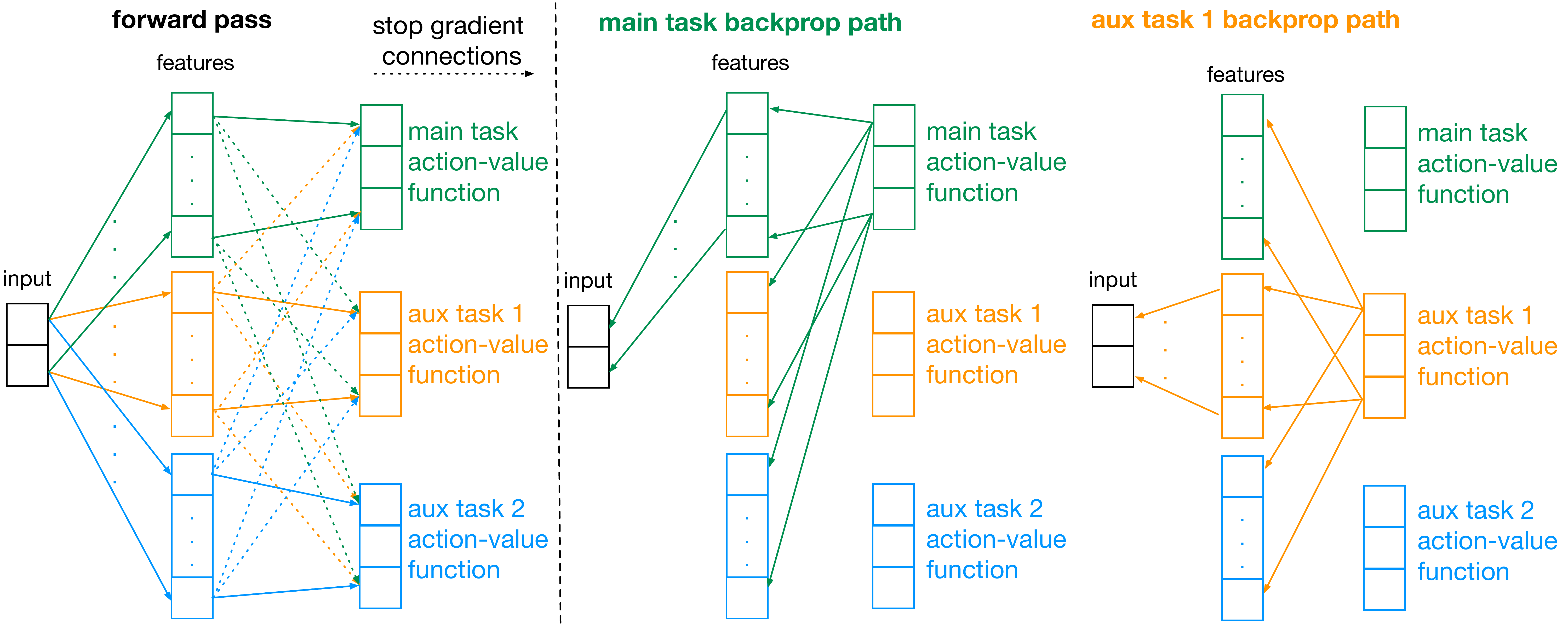}
  \caption{The forward pass, backward pass for the main task, and backward pass for auxiliary task $1$ when using the  Master-User strategy for learning auxiliary tasks alongside the main task.
  All features are used by all tasks in the forward pass but only modified through the gradient backpropagated from one task.
  The dotted arrows show stop-gradient connections. 
  The gradients does not go back any further from these connections.
  When using the Master-User strategy, it is clear which auxiliary task was responsible for inducing which feature.
  }
  \label{fig:new_learning_strategy}
\end{figure}

We propose a tester that evaluates the auxiliary tasks based on how useful the features induced by them are for the main task.
When following the standard practice of jointly learning the main task and the auxiliary tasks, recognizing which feature was influenced the most by which auxiliary task is challenging.
This is because all features are jointly shaped by all the tasks, both auxiliary and main.
To address this issue, we use a strategy for learning the representation where all features are used by all tasks in the forward pass; however, each feature is only modified through the gradient backpropagated from one task.
See Figure \ref{fig:new_learning_strategy}.
This learning strategy is similar to the Master-User algorithm proposed for continual recurrent learning \citep{javed2021scalable}.
Therefore, we refer to this learning strategy as the Master-User strategy.
When using the Master-User strategy, it is clear which auxiliary task was responsible for inducing which feature.

As we mentioned above, the proposed tester assesses the utility of an auxiliary task based on how useful the features induced by it are for the main task.
To assess each feature, the proposed tester looks at the magnitude of the outgoing weights from the feature to the main task action-value function for all actions. 
The greater the magnitude is, the more important the feature is.
The tester also considers the trace of magnitude of each feature: the greater the trace of the feature is, the more it contributes to the approximation of the main task action-value function.
The magnitude of the weights times the trace of the magnitude of the feature represents how much the feature contributes to the approximation of the main task action-value function. 
Therefore, the utility of feature $k$ is defined as:

\begin{equation}
\label{eq:1}
\begin{aligned}
u_k &= \bar{f}_k \times \sum_a |w_{ka}^{\mbox{main}}|   \\
\end{aligned}
\end{equation}

\noindent where the utility of feature $k$ is denoted by $u_k$ and $\bar{f}_k$ is a {\em trace} of feature $k$ defined as:

\begin{equation}
\begin{aligned}
\bar{f}_k &\leftarrow (1 - \tau) \bar{f}_k + \tau f_k
\end{aligned}
\end{equation}

\noindent where $f_k$ is the value of feature $k$ at the current time step and $0<\tau<1$ is the trace parameter.
This assessment method is similar to what has been used in generate-and-test on features \citep{mahmood2013representation}.

After assessing the utility of the features, the utility of each auxiliary task is set to the sum of the utility of the features shaped by it: $u(\mbox{aux}^i) = \sum_{k \in F^i} u_k$ where $F^i$ are the features shaped by auxiliary task $i$.

We combined the proposed tester with a simple generator that randomly generates auxiliary tasks.
The auxiliary task are formulated as {\em subgoal-reaching} GVFs where the continuation function returns $0$ at the subgoals and $1$ elsewhere (similar to $\gamma$ in an epsiodic MDP). 
The cumulant is $-1$ everywhere and the policy is greedy.
The subgoals are randomly selected from the observation space, meaning the agent is learning many policies to reach different parts of the observation space in addition to solving the main task.

\section{Experimental results}
In this section, we provide empirical results supporting the efficacy of the proposed generate-and-test method for auxiliary task discovery. 
We include results on two gridworld environments: four-rooms and maze.
We also include results on the pinball environment \citep{konidaris2009skill}, which is widely used in skill chaining, option discovery, and recently model-based planning \citep{lo2022goal}. We choose these environments so that we could easily visualize the discovered auxiliary tasks and easily design good and bad auxiliary task as baselines. 
All environments are episodic.

In the gridworld environments, the goal is to learn the shortest path from start state to goal. 
The start and goal states are denoted by S and G respectively in Figure \ref{fig:tester_scores} and \ref{fig:lc}. 
At each cell, four actions are available: up, down, left, and right. 
There is some degree of stochasticity when transitioning from one state to another: with probability $0.5$ the agent will transition in the same direction as the selected action, otherwise it will transition in one of the remaining directions with equal probability. 
The observation space is described with a one-hot representation with the index corresponding to the agent's position being $1$.
The reward is $-1$ on each time step.

In the pinball environment, a small ball should be navigated to the goal in a maze-like environment with simplified ball physics. 
In Figure \ref{fig:lc}, the pinball environment is shown with the ball shown by a grey circle.
The goal and start states are denoted by S and G respectively.
Collision with the obstacles causes the ball to bounce. 
The observation space is continuous and is described by $x$, $y$, $\dot{x}$, $\dot{y}$.
The start location and goal location are at $(0.8, 0.5)$ and $(0.1, 0.1)$  respectively.
The action space includes $5$ actions of increasing or decreasing $\dot{x}$ or $\dot{y}$ and no change to $\dot{x}$ and $\dot{y}$.
The reward is $-1$ at each time step. 
There is no episode cutoff.

Note that in the original pinball environment, the agent receives a special reward of $10,000$ upon arrival at the goal. Instead we gave a reward of $-1$ (like every other step) so that the scale of the action-value function for the main task and the auxiliary tasks would not be too different. 
When learning multiple tasks in parallel, the contribution of each task is determined by the scale of the corresponding value function \citep{hessel2019multi}.
Therefore, when the scale of value functions are very different, we would need to scale the reward of the main task and the cumulants of the auxiliary tasks appropriately. 
This issue requires an additional hyper parameter that would give our method an advantage if tuned. 
For this paper, we decided to focus on the case where the scale of the value function for the main task and the auxiliary tasks are similar.

We used DQN with Adam optimizer as the base learning system.
We used a neural network with one hidden layer and $tanh$ activation function.
(We used $tanh$ activation function so that the induced features would be all in the same range of $(-1, 1)$; however, our proposed tester should work well when other activation functions are used too. This can be investigated in future work.)
For the girdworld environments, the one-hot observation vector was fed to the neural network.
The hidden layer size for the baseline with no auxiliary task in four-rooms and maze were $50$ and $500$ respectively.
The replay buffer size for four-rooms and maze were $500$ and $1000$ respectively.
For both four-rooms and maze, we used a batch-size of $16$ and target network update frequency of $100$. 

For the pinball environment, the $4$-dimensional observation was normalized and fed to the neural network.
The hidden layer size was $500$.
We used a replay buffer of size $10,000$, a batch-size of 16, and target network update frequency of $200$.

\subsection{The proposed tester reasonably evaluates the auxiliary tasks}
\begin{figure}[!htbp]
\centering
  \includegraphics[width=1\linewidth]{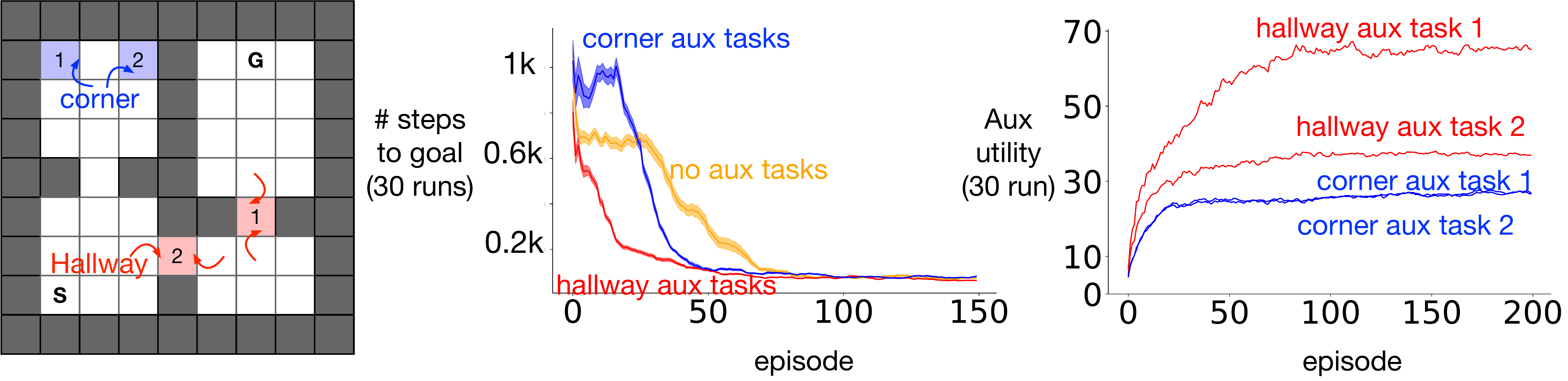}
  \caption{Left: The four-rooms environment with the subgoals corresponding to the good and bad hand-designed auxiliary tasks shown in red and blue respectively.
  Middle: Hallway auxiliary tasks improved the performance in terms of learning speed.
  The corner auxiliary tasks made learning slower in the early episodes. 
  Right: The proposed tester evaluated the hand-designed auxiliary tasks well, giving higher utility to the hallway auxiliary tasks. The results are averaged over 30 runs and the shaded regions depict the standard error.}
  \label{fig:tester_scores}
\end{figure}

To see how well the proposed tester evaluates the auxiliary tasks, we designed good and bad auxiliary tasks in the four-rooms environment.
The hand-designed auxiliary task were formulated as subgoal-reaching GVFs with the good and bad hand-designed auxiliary tasks having hallway and corner subgoals respectively.
See Figure \ref{fig:tester_scores}.
We used the Master-User architecture when learning the hand-designed auxiliary tasks. 

 When learning the auxiliary tasks alongside the main task using the Master-User strategy, the gradient backpropagated from the main task only modifies $\frac{1}{\#\text{auxiliary tasks}+1}$ percent of the features,. 
For example, in the case of learning the hallway auxiliary tasks, there are $2$ auxiliary tasks. 
Therefore, the gradient backpropagated from the main task only modifies $33.3\%$ of the features. 

When including auxiliary tasks, we adjusted the hidden layer size of the network such that the total number of learnable parameters are roughly equal across methods.
For example, in the four-rooms environment, for the case of no auxiliary tasks there are a total of $49\times 50 + 50 \times 4 = 2650$ parameters, with an observation size of (network input size) $49$ , hidden layer size of $50$, and $4$ actions (network output size). 
For the baseline with hallway auxiliary tasks, we used a smaller hidden layer size of $43$ so that it results in roughly the same number of total parameters: $49\times 43 + 43 \times 4 \times 3 = 2623$. 
In all the experiments, we kept the number of learnable parameters roughly equal across methods with and without auxiliary tasks.

The hallway auxiliary tasks improved learning in terms of learning speed as we expected (Figure \ref{fig:tester_scores}, middle graph).
The corner auxiliary tasks, on the other hand, hurt the performance at the early episodes and resulted in suboptimal performance in comparison to the hallway auxiliary tasks.
As we mentioned earlier, when including auxiliary tasks, we used a smaller hidden layer size such that the total number of learnable parameters are equal to the case of no auxiliary tasks.
It is interesting that in this environment with such a small observation space and no partial observability, dedicating a considerable percentage of the learnable parameters to learning auxiliary tasks can result in better performance compared to dedicating all the parameters to learning the main task. 

The proposed tester evaluated the hallway and corner auxiliary tasks well, assigning much higher utility to the hallway auxiliary tasks and clearly indicating the corner tasks are bad (Figure \ref{fig:tester_scores}, right graph).
The utility of all hand-designed auxiliary tasks started around the same point. 
However, the utility of the good auxiliary tasks reached a much higher level compared to the bad auxiliary tasks over time.
We also conducted a more thorough experiment in which we included all the cells as subgoals.
We observed that the tester evaluated the auxiliary tasks well, assigning higher utility to the auxiliary tasks that accelerated learning the most. 
These results are included in the appendices. 

\subsection{The generate-and-test method improves over the baseline of no auxiliary tasks}
Next, we studied the performance of the base learning system when combined with the proposed generate-and-test method.
The generate-and-test method uses the combination of the \textit{random generator} and our proposed tester.
The random generator produces subgoal-reaching auxiliary tasks with the subgoals randomly picked from the observation space. 
More specifically, in the gridworld environments, the subgoals are cells in the grid.
In the pinball environment, the subgoals are determined by $(x, y)$ and once the ball is within radius $0.035$ of a subgoal, it is assumed that the agent has reached the subgoal.

We included two baselines for comparison which included the base learning system with 1) no auxiliary tasks 2) fixed random auxiliary tasks. 
All the auxiliary tasks were in form of subgoal-reaching tasks. 
For the fixed random auxiliary tasks, the subgoals where randomly picked from the observation space and kept fixed throughout learning.
The number of auxiliary tasks for the baseline with fixed random auxiliary tasks was set equal to the number of auxiliary tasks for the generate-and-test method. 
(We also experimented with a baseline that replaces some ratio of the auxiliary tasks randomly instead of keeping them fixed. This baseline can be thought of as a generate-and-test method with a random tester and a random generator. This baseline performed worse than the baseline with fixed random auxiliary tasks. The results can be found in the appendices.)

We systematically swept the step-size parameter and report the performance of the best to ensure a fair comparison.
To do so, we ran the baseline with no auxiliary tasks with different values of the step-size for $10$ runs.
We used the step-size that resulted in the lowest final error (last $10$ percent episodes) and reran the baseline with the best step-size for $30$ runs to get the final results.
We repeated this process for all methods.
For four-rooms, maze, and pinball the sweep over the step-sizes included $\{0.000625, 0.0025, 0.01, 0.04\}$, $\{0.00025, 0.001, 0.004\}$, and $\{0.00125,0.0025, 0.005, 0.01, 0.02\}$.

The generate-and-test method has hyper-parameters of its own: 1) number of auxiliary tasks 2) age threshold 3) replacement cycle 4) replacement ratio 5) trace parameter.
For the gridworld environments, we used $5$ auxiliary tasks, age threshold of $0$, replacement cycle of $500$ steps, replacement ratio of $0.2$, and trace parameter $0.05$. 
For the pinball environment, we used $4$ auxiliary tasks, age threshold of $0$, replacement cycle of $500$ steps, replacement ratio of $0.25$, and trace parameter $0.01$.

\begin{figure}[!htbp]
\centering
  \includegraphics[width=1\linewidth]{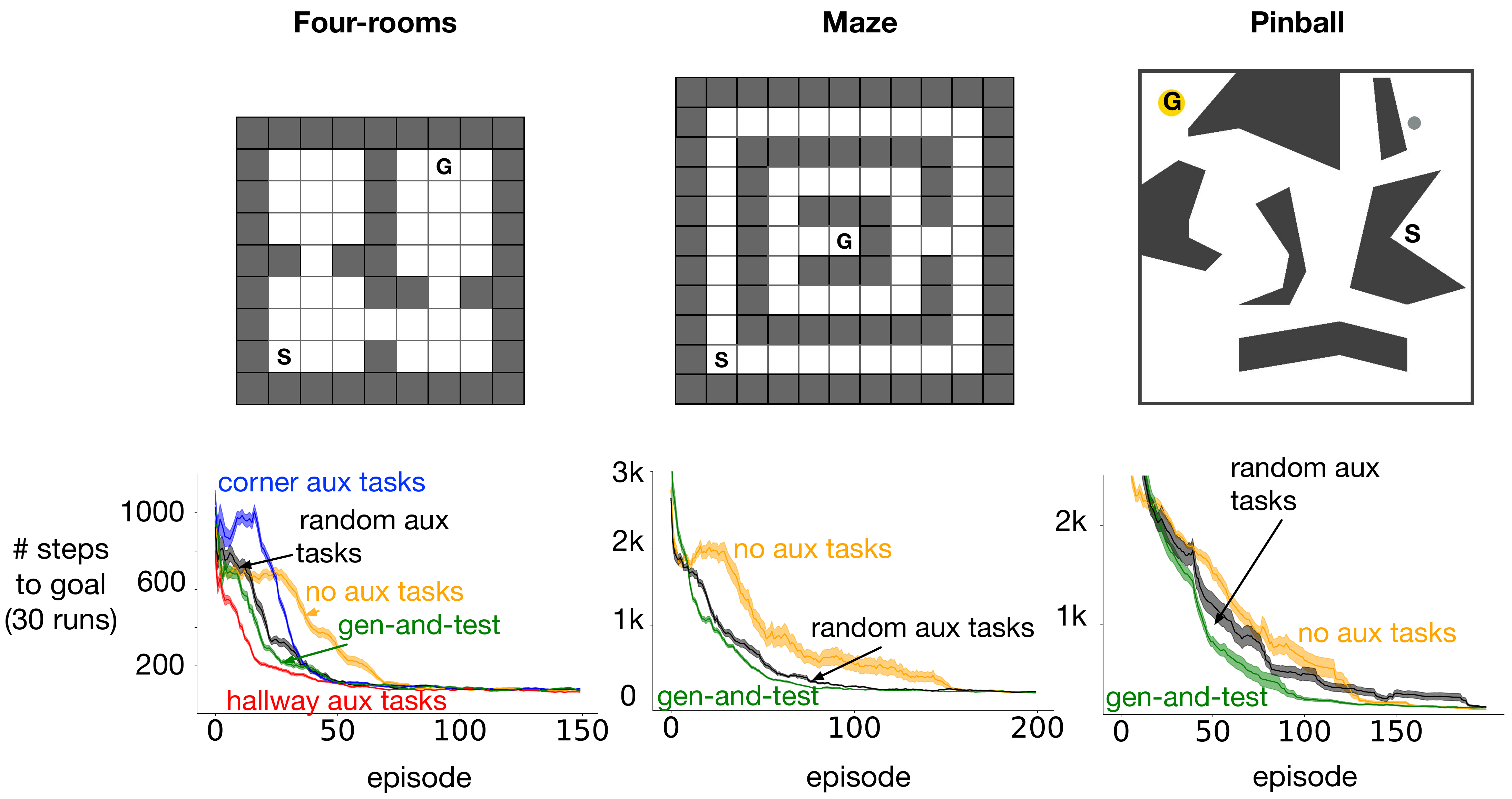}
  \caption{The learning curves for the proposed generate-and-test method (green), the baseline with no auxiliary tasks (orange), the baseline with fixed random auxiliary tasks (black). 
  The results are averaged over $30$ runs and the shaded regions depict the standard error.
  The proposed generate-and-test method improved over the baseline with no auxiliary tasks.
  Generate-and-test also outperformed the baseline with fixed random auxiliary tasks.
  Fixed random auxiliary tasks also resulted in performance gain over the baseline.}
  \label{fig:lc}
\end{figure}

First, lets take a look at the learning curves for the four-rooms environment shown in Figure \ref{fig:lc}.
The proposed generate-and-test method improved over the baseline with no auxiliary tasks bridging the gap between the baseline with no auxiliary tasks and the baseline with hallway auxiliary tasks.
Note that generate-and-test is slower than the baseline with hallway auxiliary tasks. 
This is because generate-and-test is searching the space of auxiliary tasks, starting with random ones, testing them, and retaining those recognized as useful whereas the baseline with hallway auxiliary tasks starts with reasonably good auxiliary tasks from the beginning.
The generate-and-test method outperformed the baseline with no auxiliary tasks in the maze and pinball environments as well (Figure \ref{fig:lc}).

Generate-and-test also outperformed the baseline with fixed random auxiliary tasks in all three environments.
This suggests that the choice of the auxiliary tasks was important and generate-and-test discovered and retained useful auxiliary tasks.

Interestingly, the fixed random auxiliary tasks resulted in significant performance gain over the baseline with no auxiliary tasks in the gridworld environments (Figure \ref{fig:lc}).
This is in line with the findings from the literature suggesting that random GVFs can form good auxiliary tasks for reinforcement learning \citep{zheng2021learning}.
In the pinball environment, however, the performance gain for the fixed random auxiliary tasks was small. 
This could be because the number of random auxiliary tasks is only $4$ which is relatively small for the pinball environment. 

\begin{figure}[!htbp]
\centering
  \includegraphics[width=0.9\linewidth]{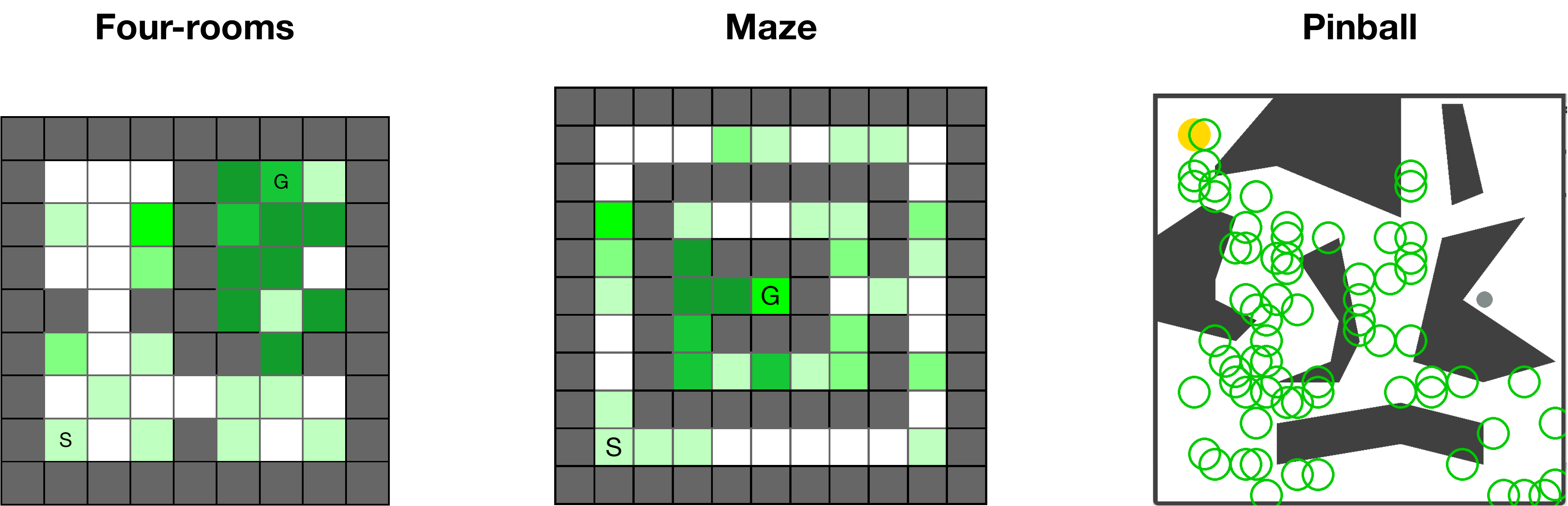}
  \caption{Example discovered auxiliary tasks in the three environments.
  Generate-and-test discovered reasonably good auxiliary tasks:
  In the gridworld environments, the subgoals corresponding to the discovered auxiliary tasks were close to the goal states. 
In the pinball environment, the discovered auxiliary tasks were more concentrated in the central areas.}
  \label{fig:discovered_goals}
\end{figure}

The auxiliary tasks discovered and retained by generate-and-test are shown in Figure \ref{fig:discovered_goals}.
To plot the discovered auxiliary tasks, we ran the generate-and-test method for 30 runs and stored the auxiliary tasks that were retained.
The green squares correspond to the discovered auxiliary tasks in the gridworld environments. 
Darker green indicates that the cell was chosen as a subgoal in many runs.
For the pinball environment, the discovered auxiliary tasks are shown by green circles. 
In the gridworld environments, the subgoals corresponding to the discovered auxiliary tasks were close to the goal states. 
In the pinball environment, the discovered auxiliary tasks were more concentrated in the central areas---reasonable way-points on the path to the goal.

\section{The feature-attainment generator}
In this section, we propose and investigate a new auxiliary task generator that has better scaling potential compared with random.
While the random subgoal-reaching generator worked well in our experiments, it may not be feasible in cases with a large observation space where the space of possible subgoals is large. 
To improve the scalability of our generate-and-test method, we propose that the discovery method searches in the space of auxiliary tasks that are about features rather than searching in the space of auxiliary tasks that are about the input observations. 

The new generator generates is based on the idea of \textit{feature-attainment} recently introduced for option discovery in planning \citep{sutton2022reward}.
In feature-attainment auxiliary tasks, the goal is to maximize an individual feature of interest (or component of the representation layer) which we refer to as the target feature.
For feature-attainment auxiliary tasks, the continuation function returns $0$ when the target feature has its maximum value and $1$ otherwise. Recall, we use $\tanh$ activation functions and thus the maximum value a feature can take on is one. 
The cumulant is $-1$ everywhere and the policy is greedy. 

When generating feature-attainment auxiliary tasks, the question is what the target features should be. 
We propose that the generator picks the target features only from the features induced by the main task. 
We designed the generator such that when picking from the features induced by the main task, it looks at the score of the features calculated according to Equation \ref{eq:1} and picks the ones with the highest scores. 
Therefore, the new generate-and-test method will be searching in the space of auxiliary tasks that are about the features that 1) contribute the most to the main task action value function 2) are induced by the main task. 

We tested the generate-and-test method with the feature-attainment generator on the three environments.
We used $3$, $8$, $4$ auxiliary tasks for four-rooms, maze and pinball respectively. 
For four-rooms and maze, we used replacement cycle of $2000$ steps, replacement ratio of $0.2$, and trace parameter $0.05$. 
For the pinball environment, we used replacement cycle of $1000$ steps, replacement ratio of $0.25$, and trace parameter $0.01$.

\begin{figure}[!htbp]
\centering
  \includegraphics[width=1\linewidth]{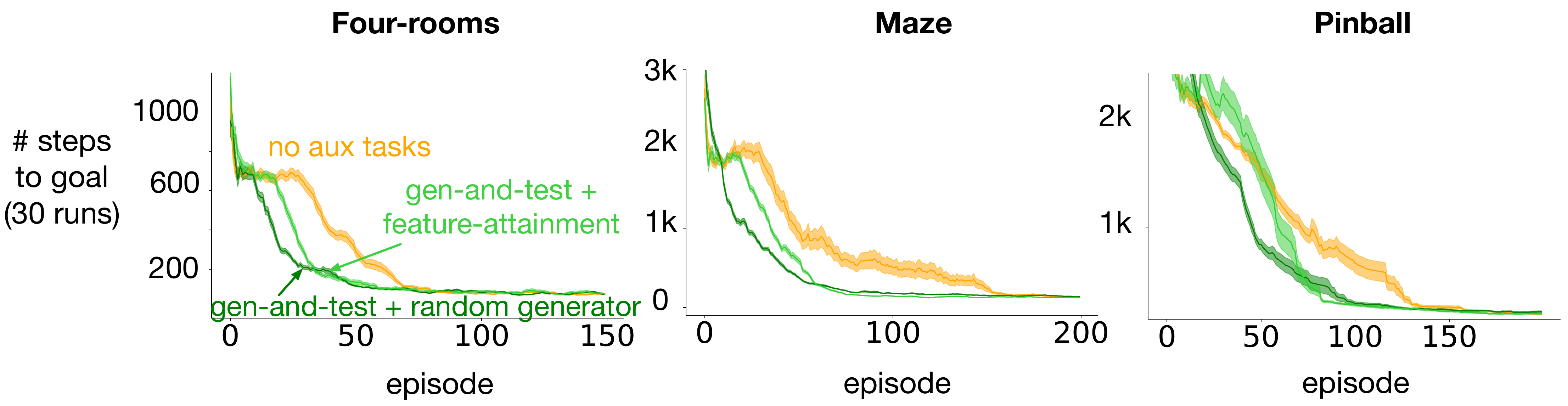}
  \caption{The learning curves for the proposed generate-and-test method with the feature-attainment generator (lime green) and the baseline with no auxiliary tasks (orange). 
  The results are averaged over $30$ runs and the shaded regions depict the standard error.
  The proposed generate-and-test method with the feature-attainment generator improved over the baseline with no auxiliary tasks and is potentially more salable than random auxiliary task generation.
  }
  \label{fig:feature_attainment}
\end{figure}

The generate-and-test method with the feature-attainment generator improved over the baseline with no auxiliary tasks across the three environments (Figure \ref{fig:feature_attainment}). 
For the first few episodes, the generate-and-test method with the feature-attainment generator did not improve over the baseline with no auxiliary tasks, unlike the variant with the random subgoal-reaching generator. 
We speculate that this is because the features are not useful in early learning and thus the feature attainment auxiliary tasks are somewhat arbitrary. With further learning, the features become more relevant to the main task progressively becoming better auxiliary tasks.

\section{Conclusions and future work}
In this paper, we proposed a new method for auxiliary task discovery.
The proposed method uses a generate-and-test approach.
We also introduced a new measure of auxiliary task usefulness.
Through careful experimentation, we showed that: 1) The proposed tester reasonably evaluates the auxiliary tasks. 2) The generate-and-test method improves over the baseline with no auxiliary tasks. 

In this paper, we took a first step toward designing a functional generate-and-test method for auxiliary task discovery.
However, there is a big space of ideas to try for designing both the generator and the tester.
For example, one idea is to sample subgoals from the replay buffer instead of generating random subgoals.
This would be similar to the idea of hindsight experience replay \citep{andrychowicz2017hindsight}.
There are also ideas from the representation learning literature to explore for improving the tester, examples of which are tested by \cite{haseeb} and \cite{elsayed2023utility}. Finally, our results with the feature attainment generator demonstrated the approach can find useful auxiliary tasks, but the next step is to try it in environments where random generation would be infeasible. We could also explore reward-respecting feature attainment subtasks \citep{sutton2022reward}, that further promote auxiliary tasks that are useful for the main task.

Another future direction is to combine the meta-learning approaches with the proposed generate-and-test method. 
The meta-learning approaches can be used to refine the auxiliary tasks discovered by generate-and-test.
It would be interesting to test the combination of meta-learning with generate-and-test in the three tested environments and see if their combination will result in discovering good auxiliary tasks, outperforming both approaches in isolation.

\bibliography{collas2023_conference}
\bibliographystyle{collas2023_conference}

\clearpage
\appendix
\section{Appendix}
\subsection{A more thorough evaluation of the proposed tester}
\begin{figure}[!htbp]
\centering
  \includegraphics[width=1\linewidth]{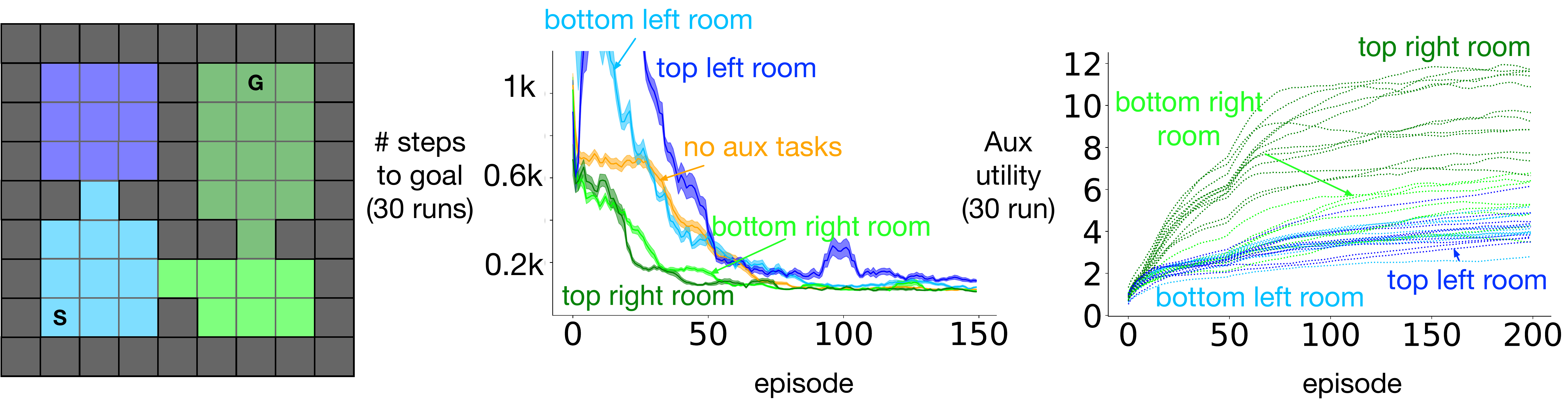}
  \caption{The tester gave higher scores to the auxiliary tasks with subgoals in the top right and bottom right rooms.
  The auxiliary tasks from the top right and bottom right rooms accelerated learning and were indeed more useful.
  }
  \label{fig:thorough_eval}
\end{figure}

In this section, we provide the result for a more thorough evaluation of the proposed tester. 
In this experiment, we included all the cells that are not a wall cell as subgoals. 
There were a total of 39 subgoals. 
We used a neural network with a hidden layer of size $240$ so each task would modify $6$ features. 
We observed that the tester gave higher scores to the auxiliary tasks with subgoals in the top right and bottom right rooms, that is subgoals closer to the goal state (Figure \ref{fig:thorough_eval}, right). 

To test whether the auxiliary tasks with subgoals closer to the goal state are indeed more useful auxiliary tasks in four-rooms, we conducted another experiment where we considered auxiliary tasks with subgoals from each room separately and compared the result with the baseline with no auxiliary tasks.
More specifically, to evaluate the subgoals from each room, in each run, we picked $5$ subgoals from that room and learned the corresponding auxiliary tasks alongside the main task. 
We did this so the number of auxiliary tasks from each room would be the same.

We observed that the auxiliary tasks from the top right and bottom right rooms accelerated learning and were indeed more useful (Figure \ref{fig:thorough_eval}, middle). 
In four-rooms, the tester evaluated the subgoals that were closer to the goal state as more useful. However, this is not necessarily the case in all environments. For example, in the pinball environment the subgoals concentrated in the central areas were recognized as more useful by the tester.

\subsection{Random tester}
In this section, we provide the result for a baseline with a random tester and a random generator. This baseline replaces some ratio of the auxiliary tasks every once in a while similar to the generate-and-test method.
However, unlike generate-and-test, it replaces auxiliary tasks at random and not based on any measure of usefulness. 
Having the auxiliary tasks replaced randomly resulted in a performance worse than that of the fixed random auxiliary tasks. 
We speculate that the poor performance of the random tester is due to the fact that randomly removing auxiliary tasks can result in the removal of useful auxiliary tasks. As a result, the overall performance might degrade.

\begin{figure}[!htbp]
\centering
  \includegraphics[width=0.4\linewidth]{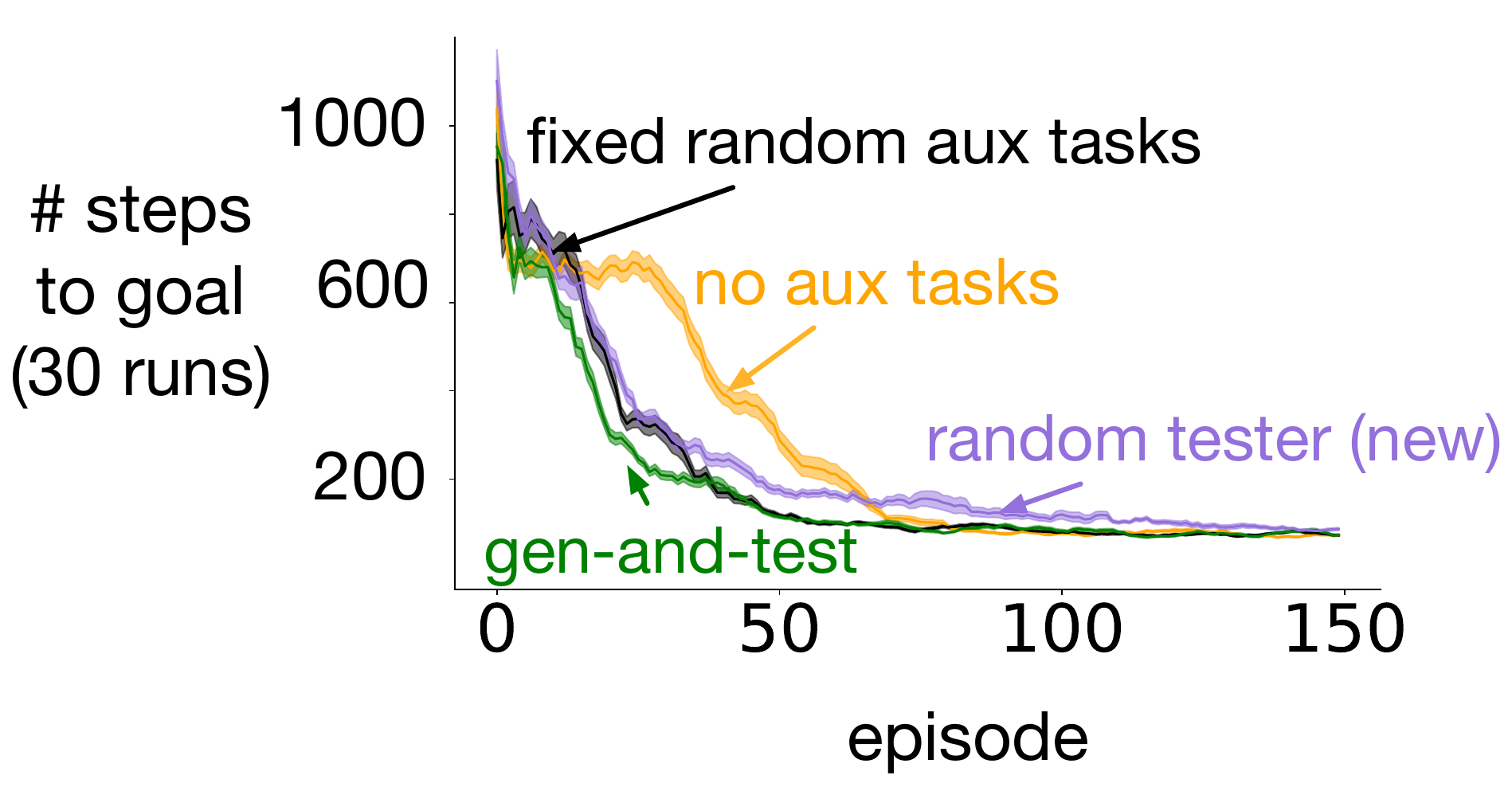}
  \caption{Having the auxiliary tasks replaced randomly resulted in a performance worse than that of the fixed random auxiliary tasks.
  }
  \label{fig:random_tester}
\end{figure}

\end{document}